\documentclass[DIV=calc, paper=a4, fontsize=11pt, onecolumn]{scrartcl}	 
\usepackage[english]{babel} 
\usepackage[protrusion=true,expansion=true]{microtype} 
\usepackage{amsmath,amsfonts,amsthm} 
\usepackage[svgnames]{xcolor} 
\usepackage[hang, small,labelfont=bf,up,textfont=it,up]{caption} 
\usepackage{booktabs} 
\usepackage{fix-cm}	 

\usepackage{times}
\usepackage{epsfig}
\usepackage{graphicx}
\usepackage{amsmath}
\usepackage{amssymb}
\usepackage{array}
\usepackage{caption}
\usepackage{xcolor}
\usepackage{authblk}
\usepackage{flushend}
\usepackage{figsize}
\usepackage{epstopdf} 

\usepackage{sectsty} 
\usepackage{fancyhdr} 
\usepackage{lastpage} 
\usepackage{lettrine} 
\usepackage{titling} 

\usepackage[margin=0.8in,letterpaper]{geometry} 
\usepackage{cite} 
\usepackage[final]{hyperref} 
\hypersetup{
	colorlinks=true,       
	linkcolor=blue,        
	citecolor=blue,        
	filecolor=magenta,     
	urlcolor=blue         
}

\allsectionsfont{\usefont{OT1}{phv}{b}{n}} 

\newcommand\eg{\emph{e.g}}

\newcommand{\HorRule}{\color{DarkGoldenrod} \rule{\linewidth}{1pt}} 

\pretitle{{ \fontsize{20}{0.5cm}\selectfont {Technical Report}} \vspace{-24pt} \begin{flushleft} \HorRule \fontsize{28}{28} \usefont{OT1}{phv}{b}{n} \color{DarkRed} \selectfont} 

\title{MICRO-EXPRESSION SPOTTING: A BENCHMARK} 

\posttitle{\par\end{flushleft}\vskip 0.5em} 

\preauthor{\begin{flushleft}\large \lineskip 0.5em \usefont{OT1}{phv}{b}{sl} \color{black}} 

\author{Xiaopeng Hong\thanks{\textit{\textit {Emails: xiaopeng.hong@oulu.fi}}}, Thuong-Khanh Tran,  Guoying Zhao}

\postauthor{\footnotesize \usefont{OT1}{phv}{m}{sl} \color{purple} 

Faculty of Information Technology and Electrical Engineering, University of Oulu, Finland

\par\end{flushleft}\HorRule} 
\date{} 

\begin{document}
\newcommand\KH[1]{\textcolor{red}{#1}}
\maketitle
\thispagestyle{empty}

\begin{abstract}
Micro-expressions are rapid and involuntary facial expressions, which indicate the suppressed or concealed emotions.
Recently, the research on automatic micro-expression (ME) spotting obtains increasing attention. 
ME spotting is a crucial step prior to further ME analysis tasks.
The spotting results can be used as important cues to assist many other human oriented tasks and thus have many potential applications. 
In this paper, by investigating existing ME spotting methods, we recognize the immediacy of standardizing the performance evaluation of micro-expression spotting methods.
To this end, we construct a micro-expression spotting benchmark (MESB). 
Firstly, we set up a sliding window based multi-scale evaluation framework.
Secondly, we introduce a series of protocols.
Thirdly, we also provide baseline results of popular methods. 
The MESB facilitates the research on ME spotting with fairer and more comprehensive evaluation and also enables to leverage the cutting-edge machine learning tools widely.
\end{abstract}


\noindent \textit{ {\color{blue}{This technical report is extended from an ACIVS17 paper. We are now expanding this work and updating this report when there are substantial achievements. The following citing information may be used for reference:}}}

\textit{ {\color{ForestGreen}{`Thuong-Khanh Tran, Xiaopeng Hong, Guoying Zhao. Sliding-window based micro-expression spotting: A benchmark. In Proc. Advanced Concepts for Intelligent Vision Systems (ACIVS), 2017'. }}}

\section{Test}

\section{Introduction}
\label{introduction}

Micro-expressions are brief involuntary facial expressions which occur when people are trying to hide true feelings or conceal emotions. 
Micro-expressions assist to understand spontaneous emotions and thus play an important role in psychology. 
Indeed, studying micro-expression facilitates to build a human behavior understanding system which can be useful in various fields such as medicine, criminal investigation, and business~\cite{li17tfc, li2015reading}. 
Therefore, this topic has been attracting increasing attention from diverse areas, such as psychology and computer science.
    
The research on micro-expression analysis can be briefly divided into two tasks: spotting and recognition, which locate positions of micro-expression in videos and determine the category of emotional states, respectively. 
To our best knowledge, most of the existing micro-expression studies focus on the recognition task \cite{polikovsky2009facial,wu2011machine,xu2016microexpression,patel16}. 
It may be because spotting is probably even more challenging than recognition. An oblique reference can be found in the human test experiments in ~\cite{li17tfc}, where the automatic ME recognition algorithm outperformed general human beings while its accuracy became slight lower than human beings when the spotting task is added to the ME analysis system.

However, for real-world applications, micro-expression positions must be determined first before further emotion recognition or interpretation. To fill in this gap, in this paper, we focus on the micro-expression spotting task. 


There are only several studies in ME spotting~\cite{li17tfc, moilanen2014spotting,patel2015spatiotemporal,xia2016spontaneous}.
Moilanen \emph{et al}. utilized the Chi-Square distance of Local Binary Pattern (LBP) to spot micro-expressions in fixed-length scanning windows~\cite{moilanen2014spotting}. 
Patel \emph{et al}. calculated optical flows for small local spatial regions, then used heuristics algorithm to filter out non-micro-expression \cite{patel2015spatiotemporal}. 
Wang \emph{et al}. suggested a method named Main Directional Maximal Differences to utilize the magnitude of maximal difference in the main direction of optical flow \cite{wang2016main}.
Basically, almost all methods focus on finding differences between non-micro and micro frames. They usually rely on  empirically-set thresholds to eliminate false alarms caused by, for example, head movement or illumination effect. 
Xia \emph{et al}. made the first attempt in utilizing machine learning for micro-expression spotting \cite{xia2016spontaneous}. Adaboost was utilized to predict whether the probability of a duration of frames belonging to a micro-expression or not. Random walk functions were used to integrate and refine the output of Adaboost and obtain the final result. 

Obviously, there are issues remaining for these ME spotting methods. 
Firstly, it's difficult to make fair comparisons between existing methods as they were usually evaluated with different test setups.
Secondly, subtle movements such as slight head movement and illumination effects are small changes occurring in continuous frames, just as micro-expressions. Thus, these extrinsic movements may lead to false alarms. 
Thirdly, existing methods usually spot micro-expression in single-scale. It probably causes problems as the length of micro-expression varies greatly. 

In addressing the first issue, it is important that experiments can be conducted under the same evaluation settings. It is thus required to standardize the performance evaluation.
For the second issue,  we tackle the spotting problem by examining a sequence of consecutive frames (which is termed a scanning window hereinafter) rather than one single frame. 
Recent state-of-the-art machine learning methods thus can be leveraged to avoid the effect of extrinsic variations.
To deal with the third issue, we introduce multi-scale analysis in micro-expression spotting.
As a result, the efforts in address these issues give rise to a benchmark of micro-expression spotting for future studies.

The contributions of this paper are three-fold:

\textbf{{Multi-scale} sliding window based approach}: {We propose a multi-scale sliding window framework.}
To our best knowledge, this is the first micro-expression spotting study combining the sliding window mechanism and multi-scale analysis. 

\textbf{Evaluation protocols}: We {standardize the comparisons of spotting methods by building} a series of protocols.

\textbf{Baseline results}:{With the proposed framework and the designed protocols, we evaluate several widely-used methods and offer baseline results for micro-expression spotting.}

This paper is organized as follows: In Section 2, proposed benchmark is detailed. In Section 3, we report the results of performance evaluation under various of parameters and test setups. Finally, Section 4 concludes our research and discusses the future works.

\section{Multi-scale Sliding-window ME Spotting}
\label{framework}

We detail the proposed benchmark including multi-scale sliding window based framework and the evaluation protocols in this Section.

\subsection{Overview of framework}
\begin{figure}
    \centering
    \includegraphics[scale=0.6]{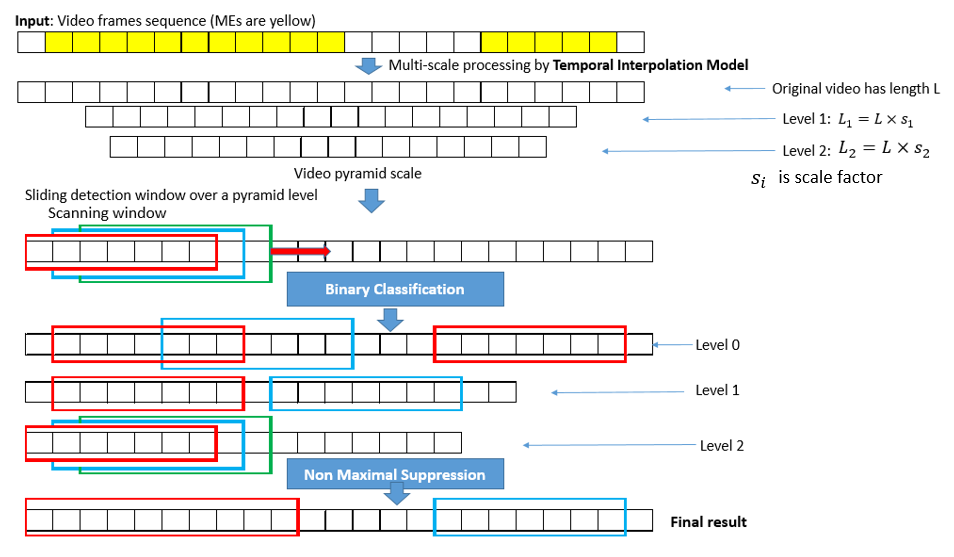}
    \caption{Illustration of sliding window based method for micro-expression spotting.}
    \label{sw_me}
     \setlength{\belowcaptionskip}{-1pt}
\end{figure}

The framework of {micro-expression spotting} has four main mechanisms as illustrated in Fig. \ref{sw_me} : 
1) multi-scale analysis, 
2) sliding window based sampling, 
3) binary classification, 
and 4) the integration and refinement of the outputs of binary classifiers.  

Firstly, we employ multi-scale analysis to rescale video sequences in temporal space, so that micro-expression of various lengths can be detected using a fix scanning window.
Temporal Interpolation Model (TIM) \cite{TIMref} has been successfully applied to ME analysis~\cite{li17tfc} and thus is utilized. 
It takes a video sequence as input and interpolates the frames to produce a sequence with a particular number of frames. 

Secondly, we sample data by using a fix-length scanning window, which is slid along all (allowed) positions of video sequences across all scales of interests to obtain samples.
These samples are labeled as micro or non-micro.
More detail will be described in the next sub-section. 

Thirdly, provided a set of samples with \emph{known} binary labels (\emph{a.k.a} the training set), it is feasible to use modern machine learning techniques such as SVM~\cite{boser1992training}, Boosting~\cite{freund1996experiments}, random forest~\cite{liaw2002classification}, or the cutting-edge deep learning methods~\cite{lecun2015deep} to optimize a binary classifier, so that the labels of \emph{unknow} samples can be predicted. 
This sub-tasks will be explained further in sub-section 3.3. 

Finally, in order to get the unique final result, it is required to integrate the binary decisions made on the densely sampled and probably overlapping scanning windows. We thus suggest a post-processing mechanism for 1) merging or integrating the nearby detected samples, and 2) removing the redundant and false positive samples. 
It can be accomplished by a temporal Non Maximal Suppression (NMS), which can be easily implemented by slightly modifying from the version for spatial domain~\cite{dalal2005histograms, redmon2016you}.



\vspace{-3pt}
\subsection{Data Sampling}

In this paper, we tackle micro-expression spotting by considering it as a binary classification task, which is carried out on a set of samples, namely fixed-length sequences of frames. 
A set of samples, especially a training set of samples with \emph{known} labels should be built first to optimize the parameters of a chosen model.
More specifically, given a video of $T$ frames, we slide a scanning window of $L$ frames to have a sample set $V$ as:

\begin{equation}
    \label{eqnsample}
    \begin{split}
    V = [A_{1} , B_{1}] \cup ... \cup [A_{i},B_{i}] \cup ... \cup [A_{M},B_{M}]
     \end{split}
\end{equation}

\noindent where $[A_{i} , B_{i}]$ is a sample starting from the frame $A_{i}$ to $B_{i}$. $A_{i+1}=A_{i} + s$ and $B_{i}=A_{i}+L-1$ for $(i=1..M)$ are indexes of the first frames and last frames of sample, respectively, and $s$ is the sliding stride. Obviously, the number of samples $M = floor(\frac{T - L}{s})+1$.

Then, by calculating the Intersection over Union of the $X_{W}$ and ground truth $X_{G}$ (as widely suggested by the object detection challenges such as the popular PASCAL VOC challenge~\cite{everingham2010pascal}), we determine the labels of samples as positive (for micro-expressions) and negative (for non micro-expressions):

\begin{equation}
    \label{overlapposneg}
    \begin{split}
    positive: \frac{X_{G} \cap  X_{W}}{X_{G} \cup  X_{W}} \geq \varepsilon
    \\ negative: \frac{X_{G} \cap  X_{W}}{X_{G} \cup  X_{W}} < \varepsilon
    \end{split}
\end{equation}

We set $\varepsilon = 0.5$ following the suggestion of~\cite{everingham2010pascal}. 

\subsection{Binary Classification}




Binary classification plays a central role in micro-expression spotting as it enables to leverage the advance in machine learning techniques for extracting the features and making binary decisions.
We follow~\cite{li17tfc} and employ three descriptors of the scanning windows  and linear support vector machine to classify the samples.

\begin{itemize}
\item Local Binary Pattern for Three Orthogonal Planes (LBP-TOP)~\cite{zhao2007dynamic}: LBP-TOP is an extension of LBP in spatial-temporal do-main. A recent efficient implementation of LBP-TOP can be found in~\cite{hong2017fast}.
\item Histogram of Oriented Gradient for Three Orthogonal Planes (HOG-TOP): HOG-TOP is extended from HOG to 3D to calculate oriented gradients on three orthogonal planes XY, XT, and YT for modeling the dynamic texture in video sequence~\cite{li17tfc}. 
\item Histogram of Image Gradient Orientation for Three Orthogonal Planes (HIGO-TOP)~\cite{li17tfc}:
HIGO-TOP is a 3D extension of Histogram of Image Gradient Orientation (HIGO). HIGO, as a degraded variant of HOG, ignores the magnitude and counts the responses of histogram bins. 

\end{itemize}

\subsection{Test Setup}
In order to standardize the evaluation, cross-validation should be performed for splitting the set of samples into a training set and a testing set. 
Two test setups, namely "normal random sampling test" and "subject independent test" are suggested. 

\textbf{Normal random sampling test}: Samples are randomly selected for the training set and the testing set. 
Division process is carried out by three suggested various splitting rates: $30\%/70\%$, $50\%/50\%$ and $70\%/30\%$. 
It is recommended to carry out 10-fold cross-validation for reliable performance evaluation. 

\textbf{Subject independent test}: In the each fold of the leave-one-subject-out setup which is previously established for ME recognition~\cite{li17tfc}, samples of one subject is used for testing and the others are for training. 

\subsection{Performance Measurement}
In this sub-section, we discuss on the measurement for comparing micro-expression detectors. According to different semantic levels, there are two measurement.

 
The first one is used to directly evaluate performance of the binary classification. It measures the accuracy on detected positive samples and negative samples. As the objects it counts are sample which are obtained by sliding windows, this measurement is named per-window measure.
 
\begin{figure}[ht]
    \centering
    \includegraphics[scale=0.6]{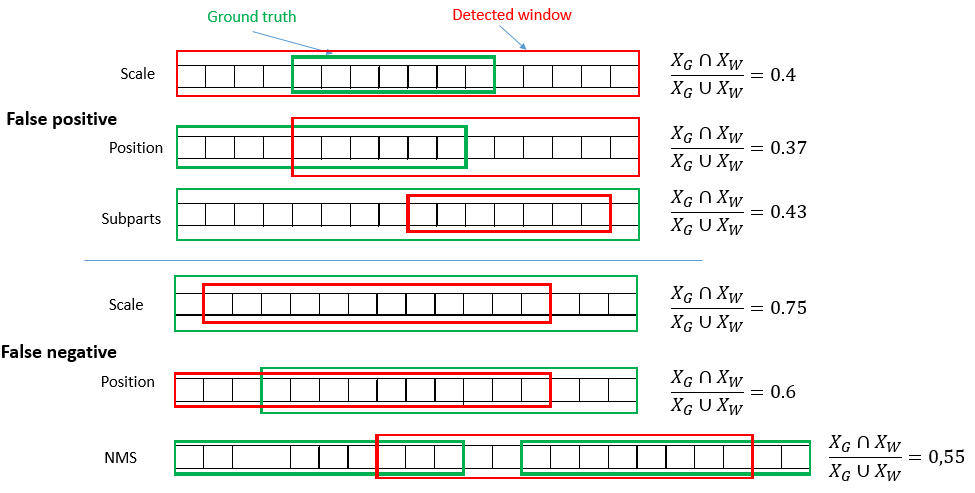}
    \caption{Untested cases of per-window evaluation. These are several cases of not testing during per-window evaluation that can lead to false positive or false negative in whole video evaluation. False positive can arise from detection at incorrect scales or positions. False negative can arise from slight mis-alignments between detected window and ground truth positions or NMS.}
    \label{pw_error}
\end{figure}

It is pointed out in the field of pedestrian detection that the per-window measure has its limitations~\cite{dollar2012pedestrian}. Recall that the target of spotting is to locate when a ME starts and ends, it is the final outputs that the performance of spotting can only be accurately evaluated by examining. However, as the per-window measure is obtained before NMS is carried out, it can only briefly reflect the performance of spotting.
Fig. \ref{pw_error} illustrates the cases which are out of the scope of per-window evaluation: multiple detected samples around ground truth, false positive samples and false negative samples caused by incorrect scales and positions. 


For these reasons, we introduce the per-video measure as the second measurement. 
Evaluation is carried out on the final outputs of detected windows, after handling the post-processing tasks and mentioned requirements. 
False positive, true positive and missing samples are determined and counted by Eq.~\ref{overlapposneg}. 

To provide a comprehensive performance evaluation of ME spotters, we utilize Detection Error Tradeoff (DET) curve. As a result, there are two types of DET curves accordingly: miss rate against false-positive per-window (FPPW) and miss rate versus false-positive per-video (FPPV). 

\section{Implementation and Results}
In this section, we provide evaluation results of selected features (HIGO-TOP, HOG-TOP, LBP-TOP) for spotting problem. Experimental results are reported by two test setups and two measurements.

\subsection{Implementation note}

We use the SMIC-VIS-E dataset  \cite{smicdata} to build the benchmark and conduct experiments. 
This dataset has 76 video sequences with the frame size $640\times480$ pixels recorded at 25fps. 
It consists of 71 micro-expression videos and 5 non-micro videos. 
These videos are annotated with onset and offset frames. 

All the frames on this dataset are pre-processed as follows~\cite{li17tfc}: face detector and KTL tracking algorithm are utilized to locate the whole faces through video \cite{viola2001rapid}. 68 land-marks points are located by Active Shape Model method \cite{cootes1995active} on the first frame of each video and they are aligned to the model face by using Local Weighted Mean (LWM) \cite{goshtasby1988image}. 
After that, face area is cropped by using the defined rectangles from the eye landmarks.

\begin{figure}[ht]
 \centering
\includegraphics[width=0.6\textwidth]{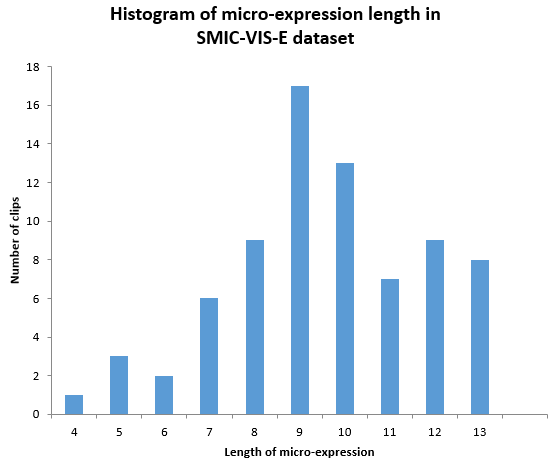}
 \caption{Histogram of micro-expression sizes in dataset.}
 \label{fig:stat1}
\end{figure}

To determine the length of the scanning windows, we count the frequency of the length of ground truths. 
Based on the plotted histogram (in Fig. \ref{fig:stat1}), the length of scanning windows is set to $9$ as it has the most number of micro-expression lengths. 
For those MEs having length other than the fixed length, we slightly modify and then fix their lengths to $9$ for simplification. 
In our experiments, $\varepsilon$ in matching between ground truth and final detected window is selected by 0.5. 


Note that a feature usually has parameters which highly affect results, thus we conduct experiments under various parameter settings to find the optimal combination. 
In our implementation, the block division is evaluated under three settings: $8\times8\times4$, $8\times8\times2$ and $6\times6\times4$. 
There are three optional overlap rates: 0.2, 0.3 and 0.5. 
The number of bins for HIGO-TOP and HOG-TOP are selected from: 8, 12 and 16. 
In Fig. \ref{fppw} and the following figures, name of descriptors explains feature used and parameter combination, \eg, \textit{HIGO-TOP bl884-ol02-nb8 means feature HIGO-TOP, block division $8\times8\times4$, overlap rate $0.2$ and number of bin is 8}. 

The evaluation experiments are conducted under both  normals random sampling and subject test setups. 

For normal random sampling test, we carry out 10 times random tests and report the means and standard deviations of the miss rates which are calculated corresponding to a series of fixed false positive rates. 
 
For subject independent test, we carry out leave one subject out cross validation. 
As the tests for random sampling, We also report results with respect to both FPPW and FPPV measure in this test. Nevertheless, there is one issue worth of mentioning for reporting results in FPPV. 
As some subject only has two videos available, there is not enough points to draw a smooth DET curve for the experiment of this fold.




 




{To address this issue,
we use Eqs.~\ref{fppv_cal} $\&$ \ref{missrate_all} for computing the (overall) FPPV and miss rate over all videos on the dataset, rather than the mean FPPV and miss rate of those on each fold of tests.}

\begin{equation}
    \label{fppv_cal}
    \begin{split}
	FPPV_{OverAll} = \frac{\sum_{j=1}^{8} FP_{j}}{N}
    \end{split}
\end{equation}

\begin{equation}
    \label{missrate_all}
    \begin{split}
	MissRate_{OverAll} = 1 - \frac{\sum_{j=1}^{8} TP_{j}}{N^{+}}
    \end{split}
\end{equation}

{ 
\noindent where $N$ is number of videos in dataset ($N=76$ in SMIC-VIS-E), $FP_{j}$ is number of false-positive in testing subject $j^{th}$, $FP_{j}$ is number of true-positive in testing subject $j^{th}$, and $N^{+}$ is the total number of ground truth MEs.
}

\begin{figure}[htbp!]
\vspace{-1pt}
\centering
\SetFigLayout{3}{2}
  \subfigure[]{\includegraphics{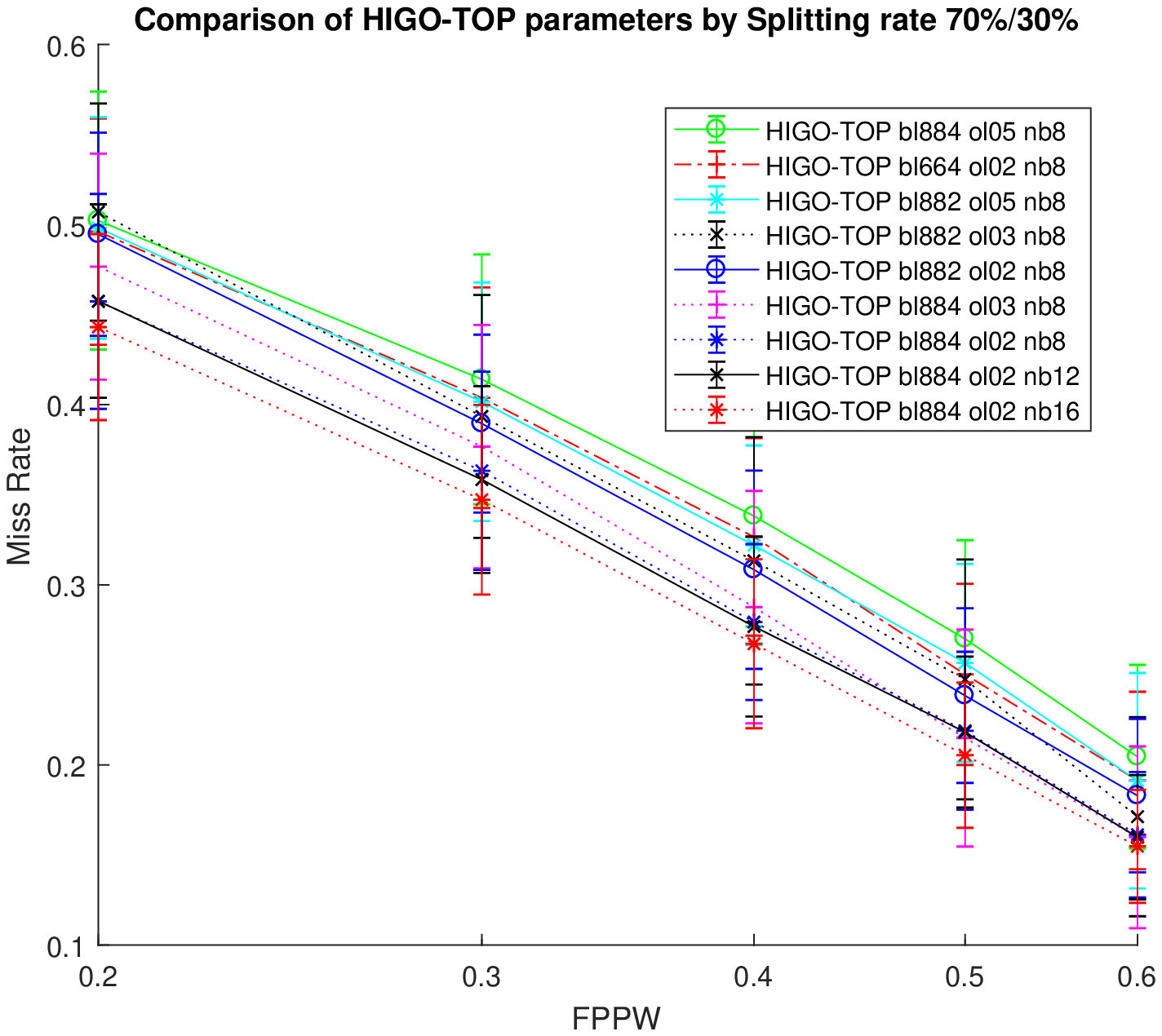}}
  \hfill
  \subfigure[]{\includegraphics{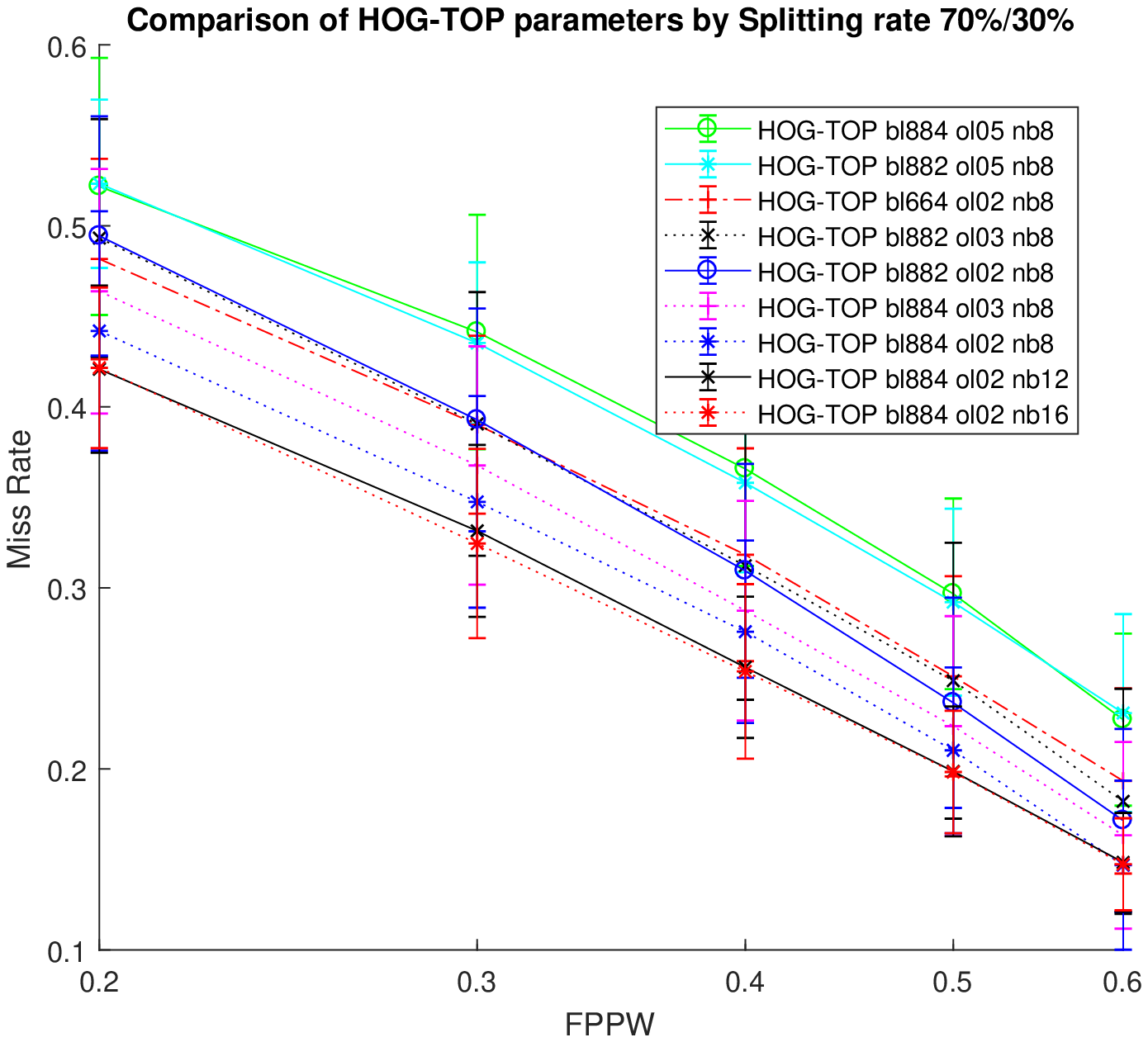}}
   \hfill
   \subfigure[]{\includegraphics{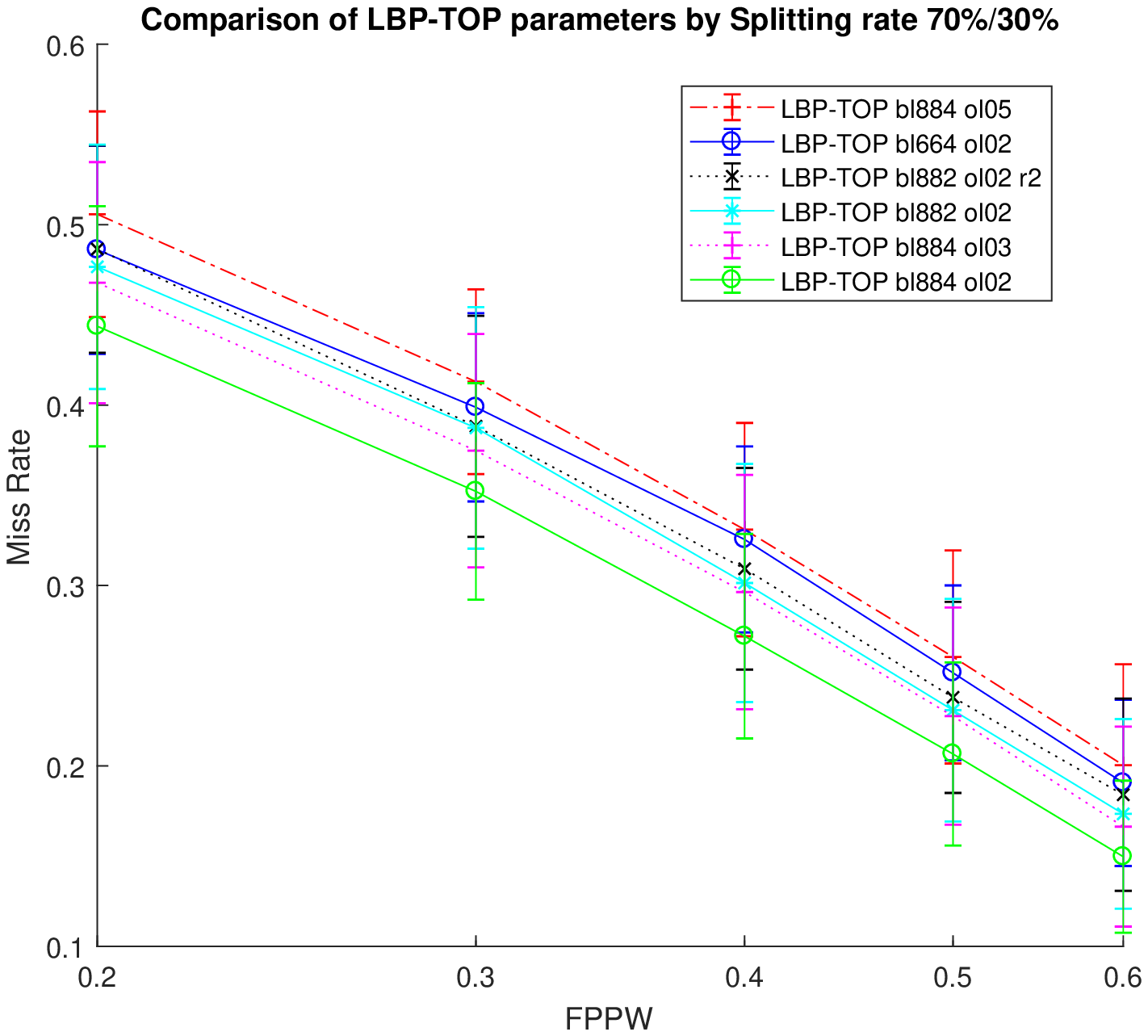}}
    \hfill
    \subfigure[]{\includegraphics{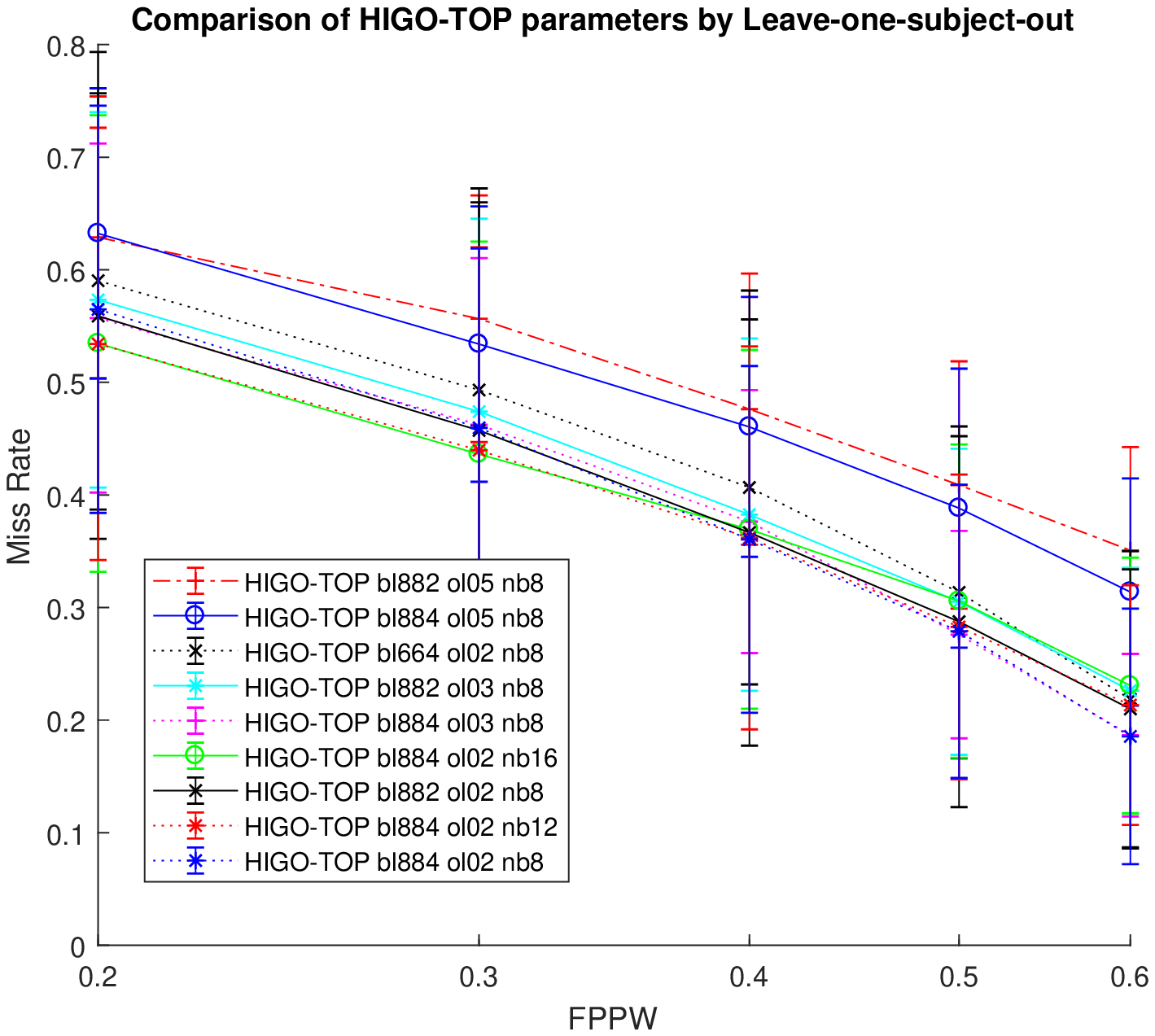}}
   \hfill
   \subfigure[]{\includegraphics{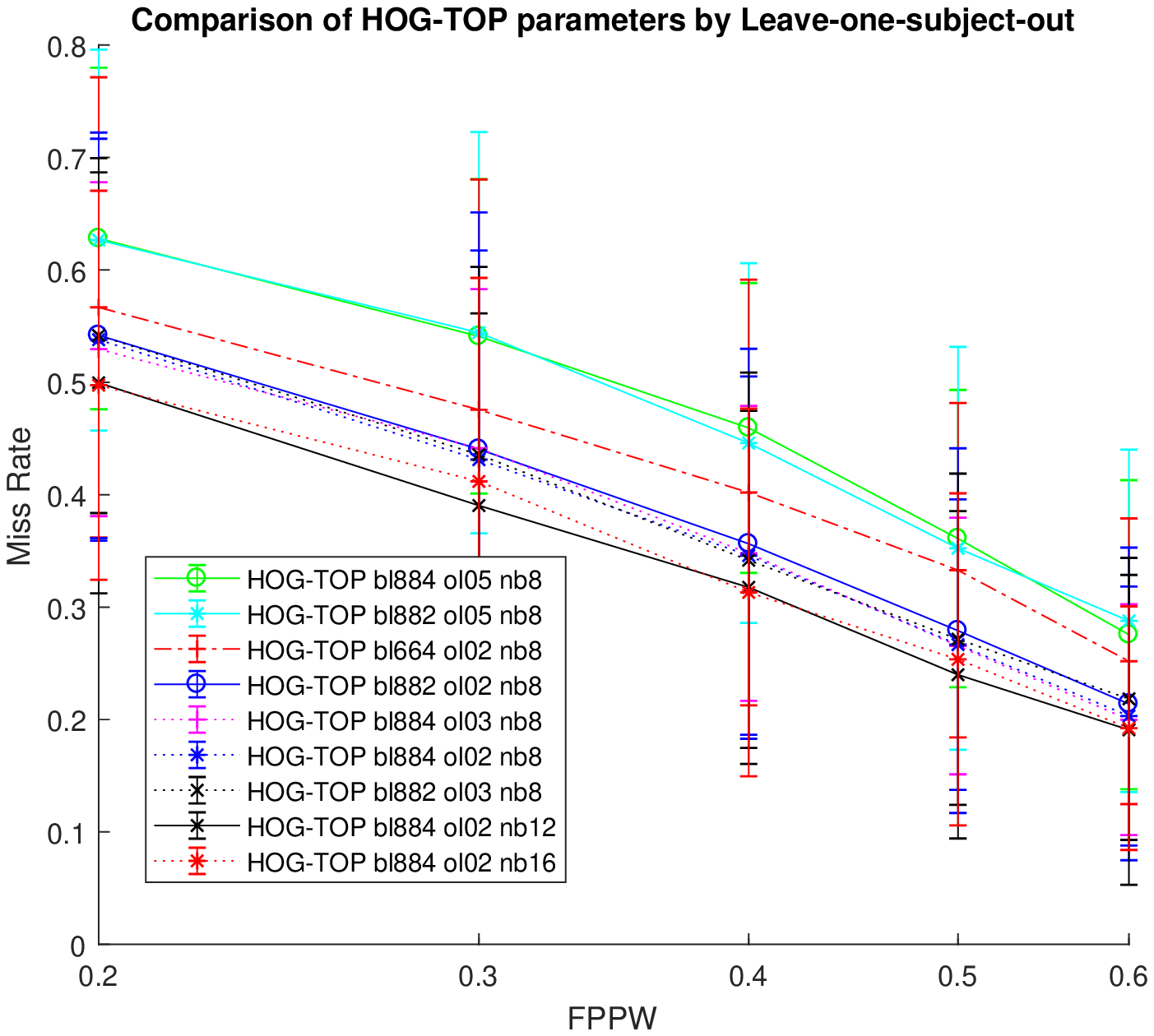}}
   \hfill
   \subfigure[]{\includegraphics{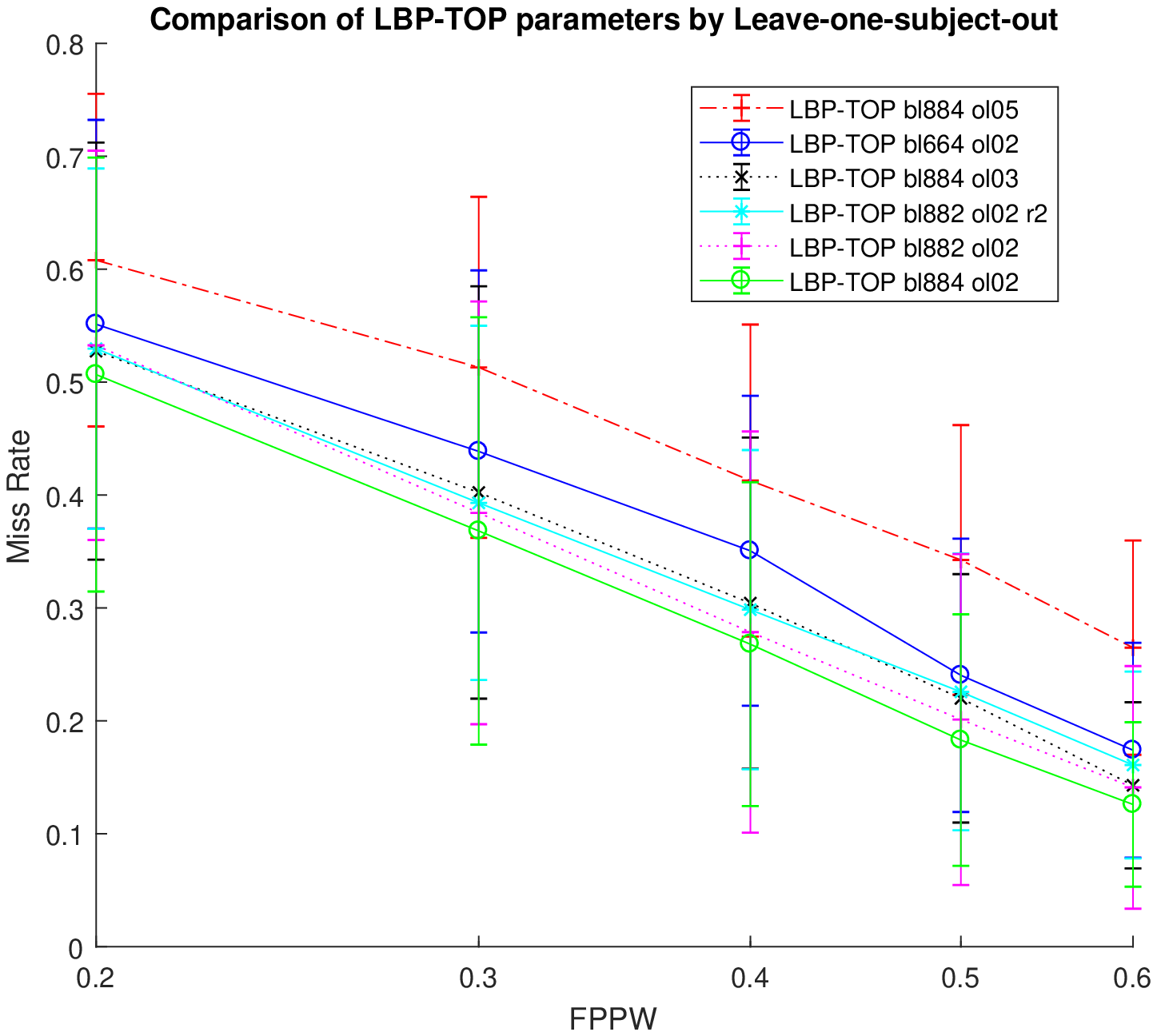}}
 \caption{{Performance of binary classifiers on per-window measurement. (a), (b) and (c) present results evaluated by normal test setup of features: HIGO-TOP, HOG-TOP and LBP-TOP, respectively. (d), (e) and (f) present results evaluated by leave-one-subject-out setup of features: HIGO-TOP, HOG-TOP and LBP-TOP, respectively. Legends are ordered by performance (lower is better).}}
 \setlength{\belowcaptionskip}{-1pt}
    \label{fppw}
 \end{figure}
 
\subsection{Per-window results}

At first, we report the results that are evaluated on per-window measurement. These results demonstrate the performance of binary classifiers and effects of parameter combinations.
For clarify, we suggest using the mean miss rates at a fixed FPPW of 0.4 as a reference points when comparing various methods.

\begin{figure}[htbp!]
\centering
\includegraphics[width=0.7\textwidth]{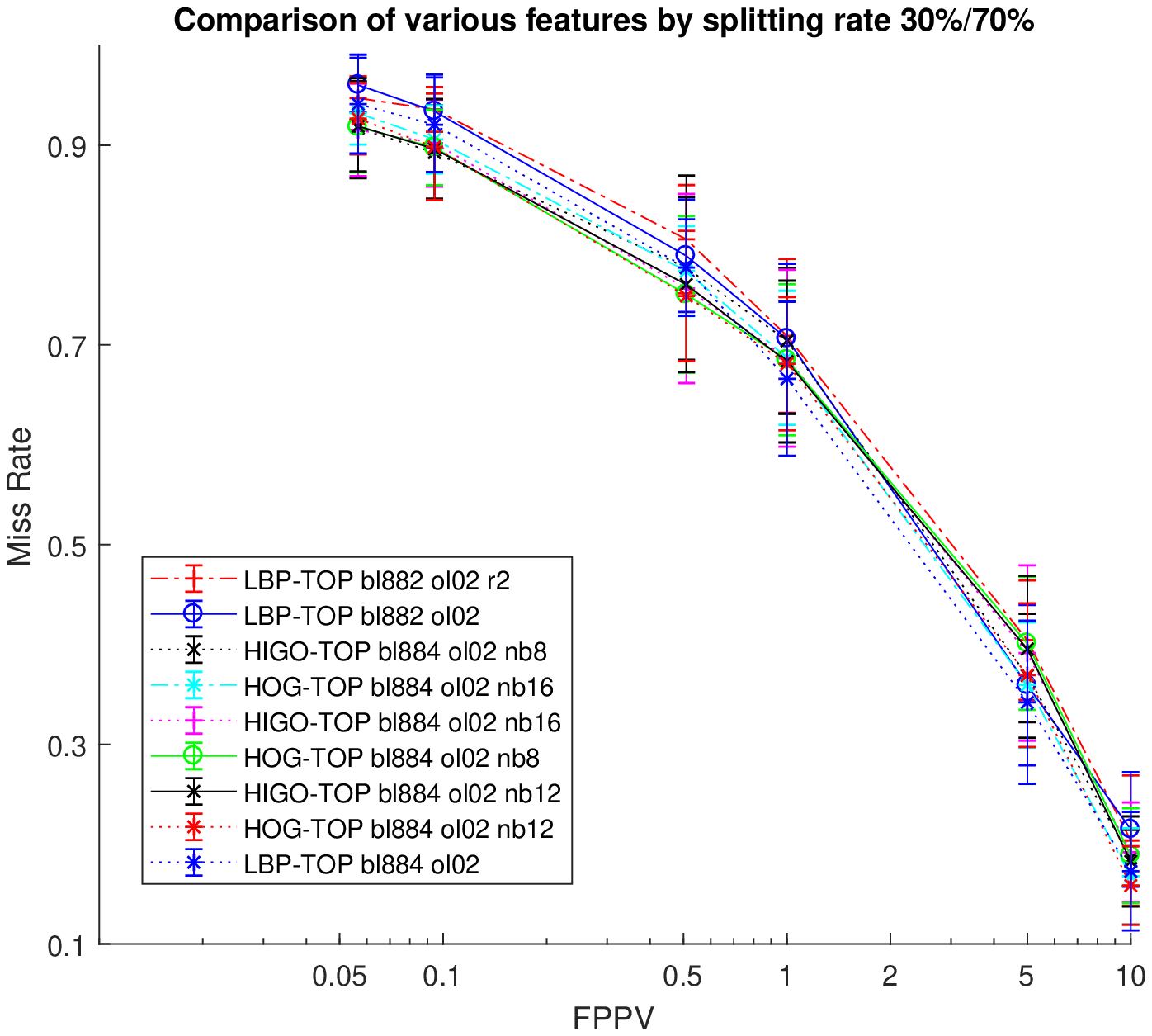}
\caption{{Evaluation results on per-video measurement, using a splitting rate of $30\%$/$70\%$.}}
\setlength{\belowcaptionskip}{-2pt}
\label{fppv-30}
\end{figure}

\begin{figure}[htbp]
\centering
\includegraphics[width=0.7\textwidth]{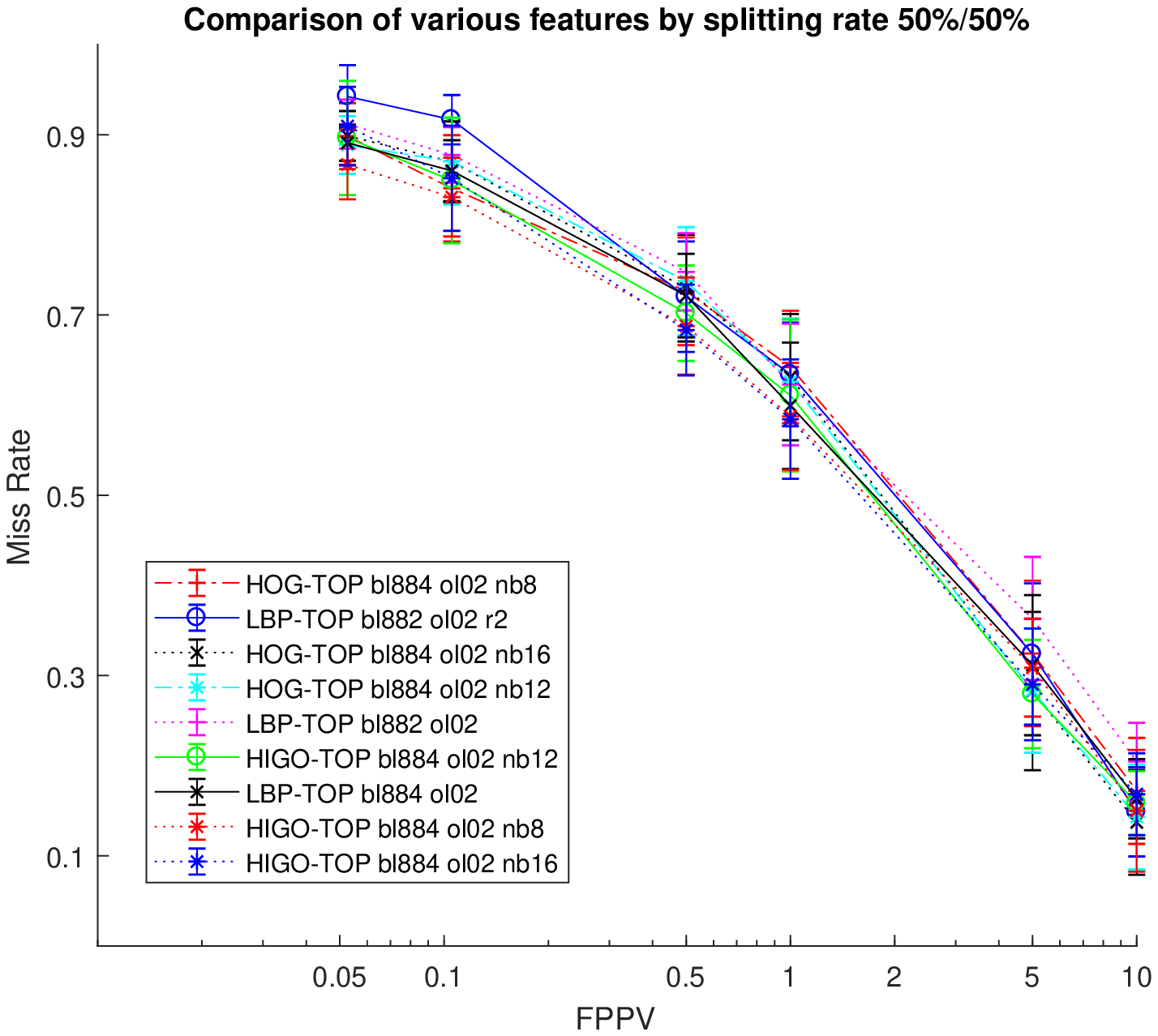}
\caption{{Evaluation results on per-video measurement, using a splitting rate of $50\%$/$50\%$.}}
\setlength{\belowcaptionskip}{-2pt}
\label{fppv-50}
\end{figure}

\begin{figure}[htbp]
\centering
\includegraphics[width=0.7\textwidth]{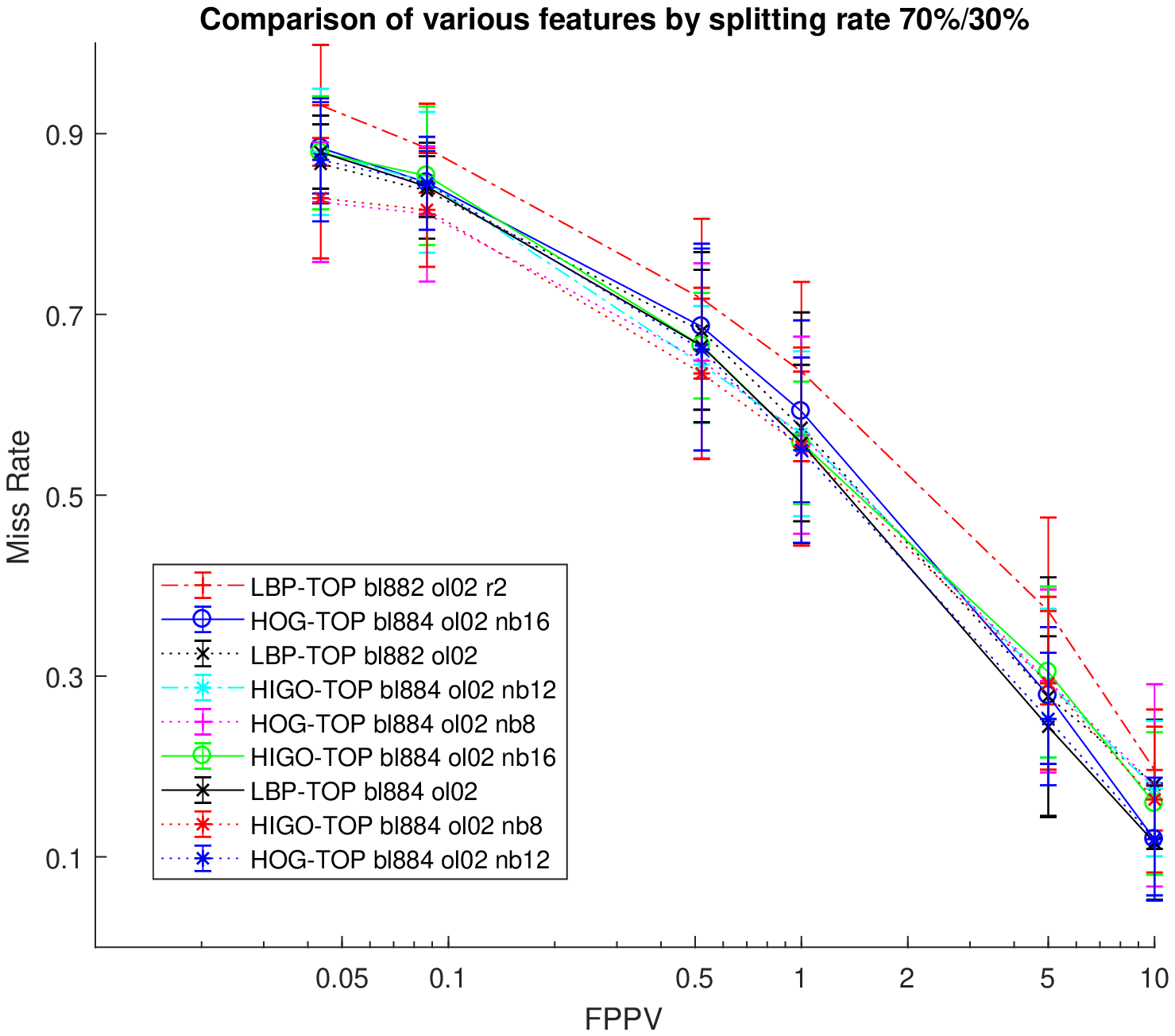}
\caption{{Evaluation results on per-video measurement, using a splitting rate of $70\%$/$30\%$.}}
\setlength{\belowcaptionskip}{-2pt}
\label{fppv-70}
\end{figure}

\begin{figure}[htbp]
\centering
\includegraphics[width=0.7\textwidth]{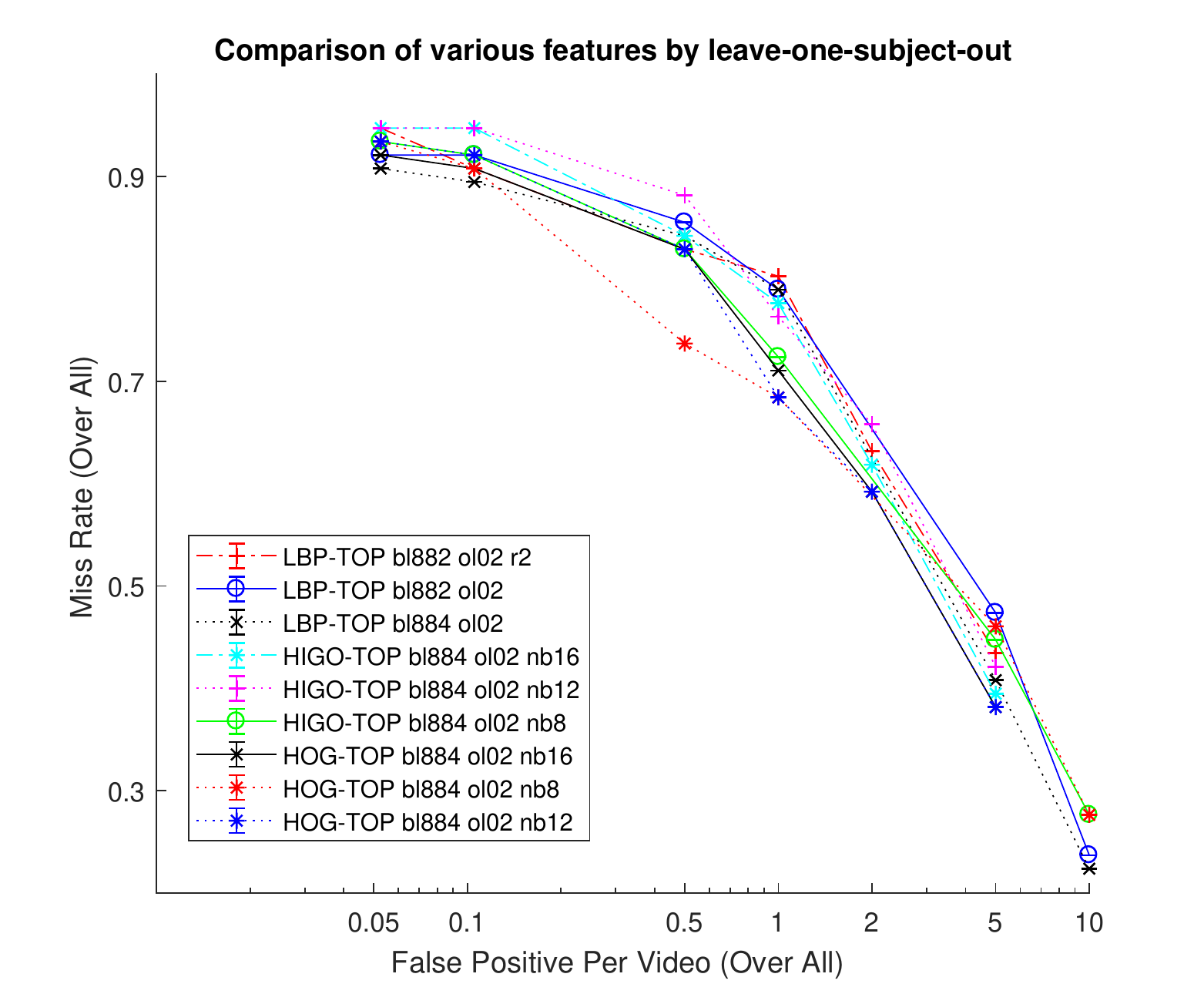}
\caption{{Evaluation results on per-video measurement, under leave-one-subject-out tests.}}
\setlength{\belowcaptionskip}{-2pt}
\label{fppv-losbj}
\end{figure}

\textbf{Normal test setup}: We discuss on the performance of classifiers evaluated by normal test setup. Fig. \ref{fppw}a, Fig. \ref{fppw}b and Fig. \ref{fppw}c plot DET curves from the performance of each feature. We present results of three features separately for comparing parameter combinations. In this setup, we only report comparison from splitting rate $70\%/30\%$ because they got the best ones. From HIGO-TOP feature, optimal block division and overlap rate are determined by $8\times8\times4$ and $0.2$. HIGO-TOP bl884-ol02-nb16 is best with mean miss rate of $26.73\%$. In HOG-TOP, HOG-TOP bl884-ol02-nb12 and HOG-TOP bl884-ol02-nb16 are the bests with mean miss rates of $25.61\%$ and $25.39.10\%$, respectively. In LBP-TOP, combination of block division $8\times8\times4$ and overlap rate 0.2 outperforms other descriptors by mean miss rate of $27.19\%$ at 0.4 FPPW. Overall, HOG-TOP is better than other features in this test setup.

\textbf{Subject independent test setup}: Next, we discuss performance of classifiers evaluated by leave-one-subject-out setup. Similar to normal test, each feature is performed separately on Fig. \ref{fppw}d, Fig. \ref{fppw}e and Fig. \ref{fppw}f. In first feature, HIGO-TOP bl884-ol02-nb12 is best one with mean miss rate of $38.47\%$ at 0.4 FPPW. In HOG-TOP, HOG-TOP bl884-ol02-nb12 performs best with mean miss rate of $33.67\%$. LBP-TOP descriptors are better than other features in this test setup. LBP-TOP bl884-ol02 and LBP-TOP bl882-ol02 obtain mean miss rates of $26.79\%$ and $27.89\%$ at 0.4 FPPW, respectively. 

\subsection{Per-video results}
In this section, we report results measured by FPPV. In these protocols, comparisons between various descriptors and effects of training/testing ratios are presented. 
As the case with respect to the per-window measure, we also use the mean miss rates at a fixed FPPV of 1.0 as a reference points when comparing various methods.

\textbf{Normal test setup}: Fig. \ref{fppv-30} provides results evaluated by splitting rate $30\%/70\%$. LBP-TOP-bl884-ol02 is best with mean miss rate of $66.61\%$ at 1 FPPV. HOG-TOP-bl884-ol02-nb12 is second with mean miss rates of $68.18\%$. In splitting rate $50\%$/$50\%$, we get the new order on performance of detectors. Fig. \ref{fppv-50} shows the evaluation results of this scenarios. HIGO-TOP-bl884-ol02-nb8 and HIGO-TOP-bl884-ol02-nb16 outperform other descriptors by mean miss rates of $58\%$. In the last splitting rate (Fig. \ref{fppv-70}), $70\%$/$30\%$, HOG-TOP-bl884-ol02-nb12 is the best one with mean miss rate of $54.99\%$ and HIGO-TOP-bl884-ol02-nb8 is second place with mean miss rate of $55.38\%$ at 1 FPPV. 

\textbf{Subject independent test setup}: In last protocol, we show the performance of detectors evaluated by subject independent test. Fig. \ref{fppv-losbj} demonstrates the results of last protocol. HOG-TOP outperforms other features in this test setup with mean miss rates of $68.42\%$ at 1 FPPV (Both HOG-TOP-bl884-ol02-nb12 and HOG-TOP-bl884-ol02-nb12 have the same mean miss rate). 

\section{Conclusion and Discussion}

In this paper, we construct a micro-expression spotting benchmark. A series of protocols and experimental settings such as the sliding window based mechanism and multi-scale analysis are introduced to standardize the evaluation. Baseline results of popular spotting methods are also provided. The experimental results strongly suggest that a saturation point of ME spotting methods is far from being reached. More cutting-edge spotting methods using, for example, Action Unit~\cite{tong2007facial} and Deep learning techniques~\cite{patel16} are therefore worth of further exploration.

\section*{Acknowledgment}

This work is sponsored by the Academy of Finland, Tekes Fidipro Program, Infotech Oulu. Moreover, Xiaopeng Hong is partly supported by the Natural Science Foundation of China under the contract No. 61572205.

\bibliographystyle{plain}
\bibliography{reference}

\end{document}